# Sentiment Analysis and Emotion Classification using Machine Learning Techniques for Nagamese Language – A Low-resource Language

**Ekha Morang[1], Surhoni A. Ngullie[1], Sashienla Longkumer[1] and Teisovi Angami[1*]**
[1]Department of Information Technology, Nagaland University, Dimapur Campus, Nagaland, India
*E-Mail: teisovi@nagalanduniversity.ac.in, teisovi@gmail.com
*Corresponding Author

**Abstract** — The Nagamese language, a.k.a Naga Pidgin, is an Assamese-lexified creole language developed primarily as a means of communication in trade between the people from Nagaland and people from Assam in the north-east India. Substantial amount of work in sentiment analysis has been done for resource-rich languages like English, Hindi, etc. However, no work has been done in Nagamese language. To the best of our knowledge, this is the first attempt on sentiment analysis and emotion classification for the Nagamese Language. The aim of this work is to detect sentiments in terms of polarity (positive, negative and neutral) and basic emotions contained in textual content of Nagamese language. We build sentiment polarity lexicon of 1,195 nagamese words and use these to build features along with additional features for supervised machine learning techniques using Naïve Bayes and Support Vector Machines.

## 1. INTRODUCTION

Nagamese (*Naga Pidgin*) is an important Creole language, apart from the tribal languages spoken which is used as a common language across the entire state of Nagaland which is located in the North-Eastern part of India. It is an Assamese-lexified (Assam is an Indian state bordering Nagaland) creole language developed primarily as a means of communication in trade between the people of Nagaland and people from Assam. It is widely used in mass media in the news, radio stations, state-government media, etc. The Nagamese language is a resource-poor language, and therefore, it is a challenge to create resources for applying various language processing tasks.

Sentiment Analysis is a natural language processing (NLP) task that deals with capturing the opinion or sentiment of the person towards something such as an object, event, etc. The target of Sentiment Analysis is to find these opinions, identify the sentiments expressed, and then classify them in terms of their polarity such as positive sentiment, negative sentiment or, neutral sentiment. Identifying the emotion of a person is the task in which emotions such as anger, joy, etc., are captured. These tasks are useful in building applications such as predicting market trend based on sentiment analysis, prediction of election results based on emotion distribution, etc.

The aim of this work is to identify the sentiment and emotion contained in Nagamese textual content. The objectives of this research are:

- Classification of sentiments in terms of polarity (*positive, negative, and neutral*) in the textual content of Nagamese language.

- Classification of basic emotions *(anger, anticipation, disgust, fear, joy, sadness, surprise,* and *trust)* contained in the textual content of Nagamese language.

- Building sentiment polarity lexicons and using this to build features along with additional features for supervised machine learning techniques using Naïve Bayes and Support Vector Machines.

The rest of the paper is organized as follows: Section 2 gives an introduction to the Nagamese Language, Section 3 gives an overview of the related works, Section 4 describes the dataset creation, Section 5 discusses the methodology in detail, Section 6 provides details of the experimental setup, Section 7 discusses the experimental results and Section 8 draws the conclusion and discusses future works

## 2. THE NAGAMESE LANGUAGE

This section provides an overview of the character set, syllabic pattern, grammar, lexical, and other resources for the Nagamese Language.

### 2.1 Character Set for Nagamese Language

The Nagamese language has 28 phonemes, including 6 vowels and 22 consonants (Sreedhar 1985).

Vowels: i, u, e, ə, o, a

Consonants: p, t, c, k, b, d, j, g, $p^h$, $t^h$, $c^h$, $k^h$, m, n, ń, s, š, h, r, l, w, y

An example sentence in Nagamese is given below:

*'Moy dos baje pora yeti ase.' (I am waiting here from 10 o'clock.)*

### 2.2 Syllabic Pattern

A Nagamese word may consist of one or more syllables ranging upto a maximum of four syllables (Sreedhar 1985). The entire monosyllabic words can be subgrouped into six classes as given in the formula:

$$(C)\ (C)\ V\ (C)\ (C)^2$$

The only limitation in the above formula is that V cannot occur alone.

Nagamese exhibits disyllabic, trisyllabic, and tetra syllalabic. There are no pentasyllabic words unless one takes clear compound words.

## 2.3 Grammar

Nagamese is a Subject-Object-Verb (SOV) (Singh 2021) language similar to other Indian languages such as Hindi, Assamese, etc. The detailed grammar of Nagamese can be found in the works of Sreedhar (1985), Baishya (2003), and Bhattacharjya (2003).

## 2.4 Lexical and Other Resources

Listed below are some of the lexical and other resources that are available for Nagamese Language.

**Table 1: Nagamese resources**

| Resource name | Remarks |
|---|---|
| The Dictionary of Nagamese Language: Nagamese-English-Assamese by Bhim Kanta Boruah. | This Dictionary contains the lexical category (POS information) of every word. |
| https://nagamesekhobor.com/ | Contains news articles in the Nagamese Language. |
| Nagamese Bible (BSI) | Can be accessed online from the weblink https://www.bible.com/bible/3335/MAT.1.ISVNAG |
| Books, Hymns, etc written in Nagamese | - |
| Social Media contents in facebook, twitter, etc | - |

# 3. RELATED WORKS

Since Nagamese is an Assamese-lexified creole language, we present here some of the works that have been done for the Assamese language.

Das et al. (2021) built a sentiment polarity classification model by applying machine learning classifiers for Assamese language using lexical features like adjectives, adverbs, and verbs on the news domain.

Dev et al. (2021) attempt to perform Sentiment Analysis on Assamese Texts using the concepts of popular sentiment analyzer named “Vader”, while taking “Bengali-Vader” as its backbone.

Das et al. (2022) presented an Assamese multimodal dataset from the news domain and proposed a multi-stage multimodal sentiment analysis framework that uses textual and visual cues to determine the sentiment.

Das et al. (2023) introduced an image–text multimodal sentiment analysis framework, i.e., Textual Visual Multimodal Fusion for Assamese. In their work, they also built an Assamese news articles dataset consisting of news text and associated images and one image caption to conduct an experimental study.

Dev et al. (2023a) investigated sentiment classification on Assamese textual data using a dataset created by translating Bengali resources into Assamese using Google Translator. The work employs supervised ML methods, such as Decision Tree, K-nearest neighbour, Multinomial Naive Bayes, Logistic Regression, and Support Vector Machine, etc.

Dev et al. (2023b) proposed a deep neural network hybrid model, combining Convolutional Neural Network (CNN) and Long Short-Term Memory (LSTM), termed LSTM-CNN, in predicting sentiment polarity for Assamese.

Das et al. (2024) proposed a hybrid fusion-based multimodal sentiment analysis framework, i.e., textual, visual, and multimodal systems for the Assamese news domain. They used lexical features and specific image objects, creating a feature-level fusion to introduce multimodal for joint sentiment classification.

## 4. DATASET CREATION

To apply any learning model, we need to create a training dataset to train the learning model. This section describes the details of the annotated sentiment polarity and the emotion dataset. The unlabelled dataset containing approx. 25k tokens and 1,195 unique words was created by collecting articles from online Nagamese News articles (https://nagamesekhobor.com/ ) and bible passages (https://www.bible.com/). The Nagamese Khobor contains a variety of content related to current state affairs, sports etc. Random news articles from Nagamese khobor were collected from which various articles were extracted in order to obtain a mixed corpus of 594 Nagamese sentences. This was equally distributed among annotators of 198 sentences each. Fig. 1 shows the word cloud for the Nagamese raw dataset based on the word frequency in the unlabelled dataset. Fig. 2 shows the sample of labelling dataset. Fig. 3 provides the polarity percentage in the labelled dataset. And Fig. 4 provides the emotion distribution of the labelled dataset.

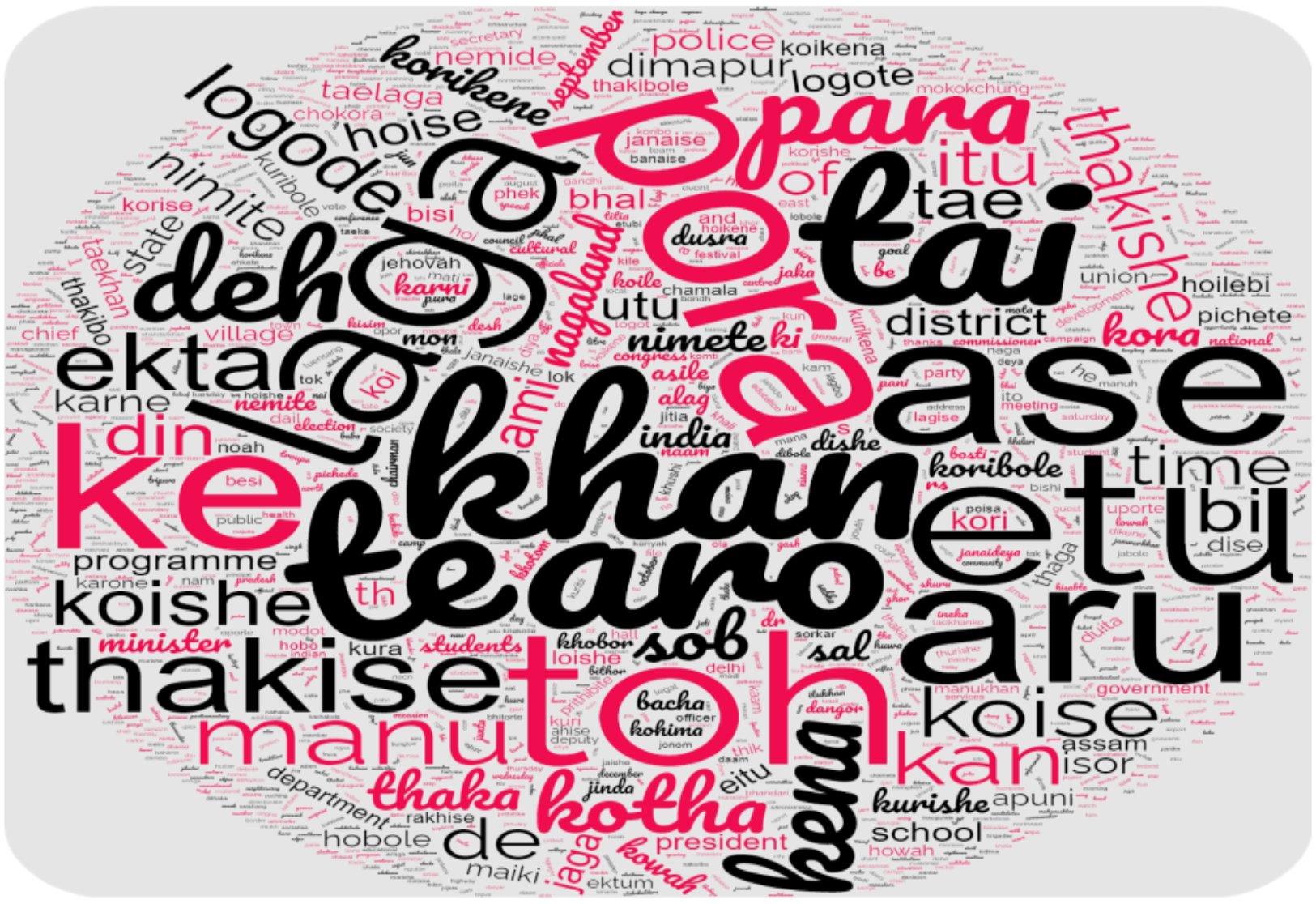

**Fig. 1: Raw dataset word frequency**

| | | |
|---|---|---|
| **1** | Titia Isor koise,"Ujala hobole dibi."Aru Ujala hoise. | Joy |
| **2** | Ineka pora prithibite sob kisim laga ghaskhan aru ghaskhan hobole lagise, bijon dhuria ghashkhanbi hobole lagise aru itu dikhikene Isor khusi lagise. | Joy |
| **3** | Itukhan prithibike ujala koribole nimite akashte chomkibo. "Aru thiknineka he hoise. | Anticipation |
| **4** | Noah hosate thakai ekta manu asile, jun taelaga homoete khali ekjon he ase jun Isor laga kotha manikene thake. | Trust |

**Fig. 2: Sample annotated dataset**

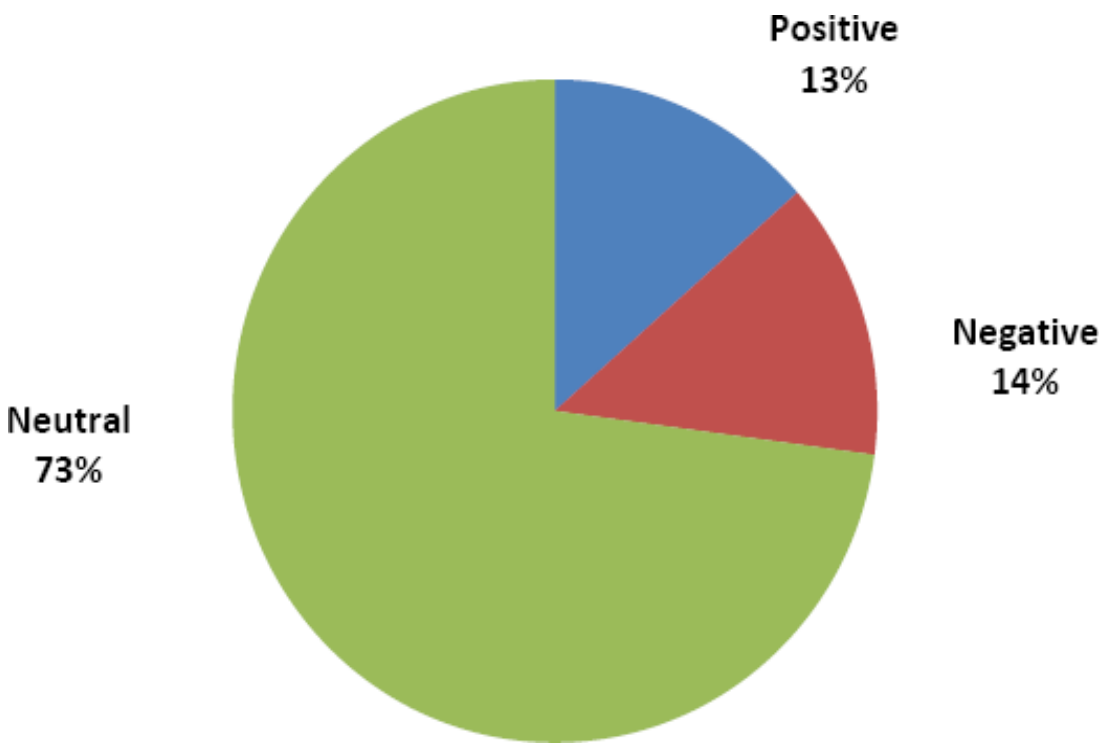

**Fig. 3: Polarity percentage**

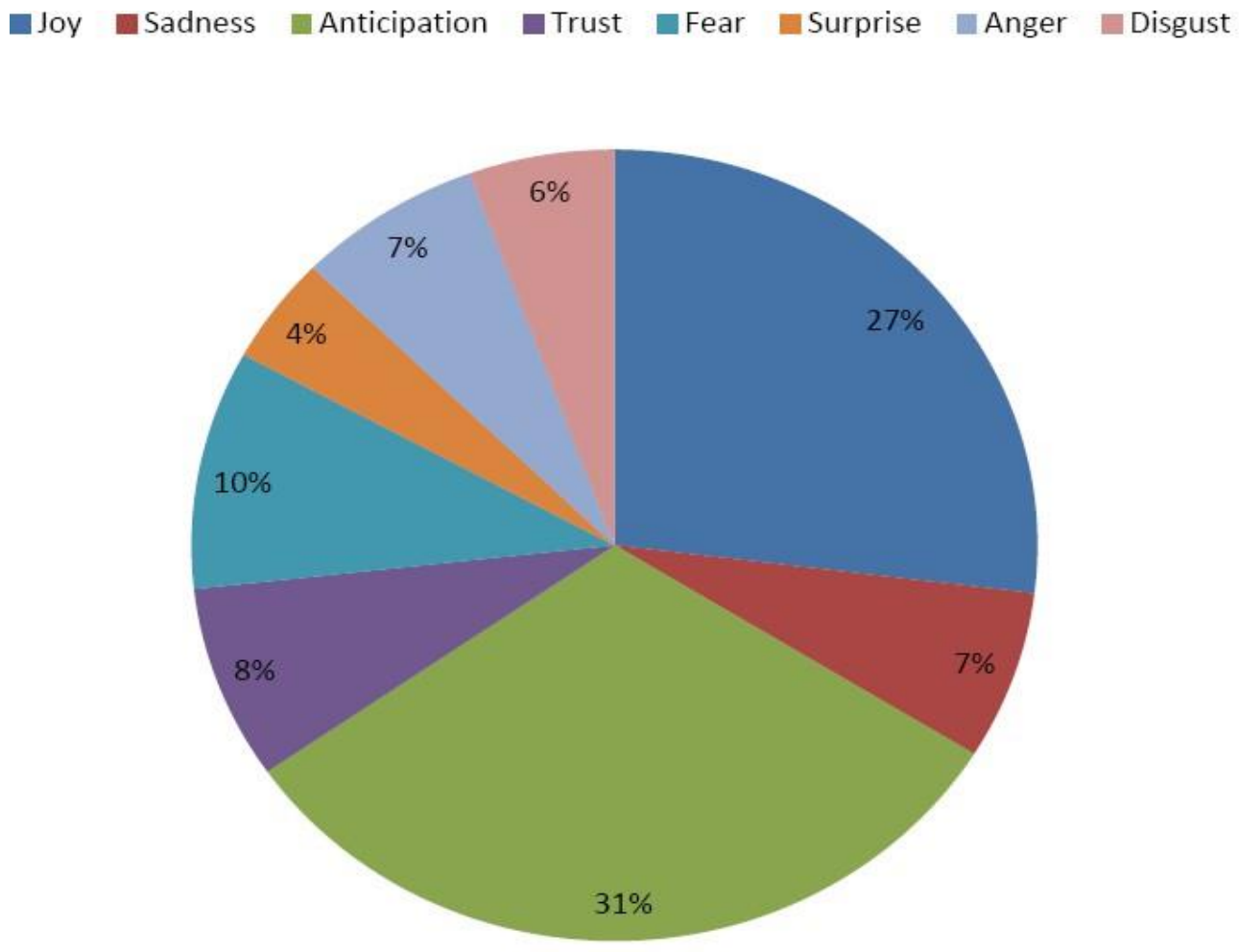

**Fig. 4: Emotion distribution**

## 5. METHODOLOGY

In our methodology, we build a sentiment dictionary of 1,195 Nagamese words which is used to generate additional features for the classification purpose. This section provides details of the sentiment dictionary, which has been used to build features for applying the machine learning models, along with other features.

### 5.1 Sentiment Dictionary

From the unique words list collected, we categorized them into positive, negative, and neutral. Table 2 provides the details of the sentiment dictionary. Table 3 shows some example words in the Sentiment dictionary.

**Table 2: Sentiment dictionary**

| Sentiment | No. of words |
|---|---|
| Positive | 162 |
| Negative | 162 |
| Neutral | 871 |

**Table 3: Sentiment dictionary example words**

| Sentiment | words |
|---|---|
| Positive | aramse, asis, bachabole, laap, modop, ujala, chanbin, bhal, biswas, hosiyar, jeetibo,... |
| Negative | bothnam, bhoe, biya, chamala, diktar, gali, galdi, hamlaa, karikena, jokhom, larai,... |
| Neutral | aage, ahibo, anibo, ase, banaisile, bhabesi, bhitor, bosti, dikheshe, ghumaise, that,... |

### 5.2 Feature generation

For the purpose of creating features for the machine learning models, we selected the features based on the sentiment dictionary and other features. We generate the count of positive, negative, and neutral words using the sentiment dictionary that we built. We used the same set of features for sentiment polarity and emotion classification. The complete list of features used is mentioned below.

Features for prediction:
1. Length of the sentence.
2. Count of positive words.
3. Count of negative words.
4. Count of neutral words.
5. Count of positive intensity words.
6. Count of negative intensity words.
7. Occurrence of the word “bisi” (A lot/large).
8. Occurrence of the word “olop” (small/less).
9. Occurrence of the emoticon ☺.
10. Occurrence of the emoticon ☹.
11. Occurrence of the exclamation mark.
12. Occurrence of the question mark.

From the full list of features, we selected the best features for the classification as given below.

Best features set for prediction:
1. Count of positive words.
2. Count of negative words.
3. Count of positive intensity words.
4. Count of negative intensity words.
5. Occurrence of the word “bisi” (A lot/large).
6. Occurrence of the word “olop” (small/less).
7. Occurrence of the emoticon ☺.
8. Occurrence of the emoticon ☹.
9. Occurrence of the exclamation mark.

## 6. EXPERIMENTAL SETUP

For our experiments, we have chosen the *Naïve Bayes* Webb (2010) and *Support Vector Machines* (SVM) Hearst (1998) Machine Learning Classifiers. The class labels for Sentence polarity are *positive, negative,* and *neutral*; and emotions are *anger, anticipation, disgust, fear, joy, sadness, surprise,* and *trust*.

To evaluate our learning models, we split the annotated dataset into 494 and 100 sentences for training and testing, respectively, as shown in Table 4.

**Table 4: Training and test dataset**

| Particulars | No. of sentences |
|---|---|
| Training | 494 |
| Test | 100 |
| Total | 594 |

For implementation, we used *scikit-learn* (https://scikit-learn.org/), which is an open-source machine learning library for the Python programming language. For Naïve Bayes, we import GaussianNB class. For SVM, we import the SVC class. We perform the SVM experiments for linear, poly and rbf kernels, and the best result is reported. Other parameter settings for the SVC are shown below:

SVC(C=1.0, cache_size=200, class_weight=None, coef0=0.0, decision_function_shape='ovr', degree=3, gamma='auto', kernel='rbf', max_iter=-1, probability=False, random_state=None, shrinking=True, tol=0.001, verbose=False)

## 7. EXPERIMENTAL RESULTS & DISCUSSIONS

This section reports the sentiment polarity and emotion classification results for the machine learning techniques. Table 5 reports the performance evaluation of sentiment classification for the experiments that we conducted. The accuracy is calculated using the *accuracy_score* class from *sklearn.metrics*. We obtain the highest accuracy for both sentiment and emotion using SVM classifier, with a score of 71% using *rbf* kernel and 67% respectively using *poly* kernel. Table 6 shows the SVM polarity classification accuracy details in terms of precision, recall and f1-score for each sentiment class. We obtained f1-score of 0.58 for negative, 0.69 for neutral, and 0.76 for positive. Table 7 shows the SVM emotion classification accuracy details in terms of precision, recall, and f1-score for each emotion class. We obtained the highest f1-score of 1.00 for fear and surprise, and least of 0.00 for anger and trust.

**Table 5: Performance evaluation of sentiment classification**

| ML technique | sentiment accuracy | emotion accuracy |
|---|---|---|
| Naive Bayes | 59% | 21% |
| SVM | **71%** | **67%** |

**Table 6: SVM polarity classification accuracy details**

| polarity | precision | recall | f1-score |
|---|---|---|---|
| negative | 0.64 | 0.54 | 0.58 |
| neutral | 0.70 | 0.68 | 0.69 |
| positive | 0.73 | 0.78 | **0.76** |

**Table 7: SVM emotion classification accuracy details**

| emotion | precision | recall | f1-score |
|---|---|---|---|
| anger | 0.00 | 0.00 | 0.00 |
| anticipation | 0.68 | 0.67 | 0.68 |
| disgust | 1.00 | 0.50 | 0.67 |
| fear | 1.00 | 1.00 | **1.00** |
| joy | 0.62 | 0.78 | 0.69 |
| sadness | 1.00 | 0.67 | 0.80 |
| surprise | 1.00 | 1.00 | **1.00** |
| trust | 0.00 | 0.00 | 0.00 |

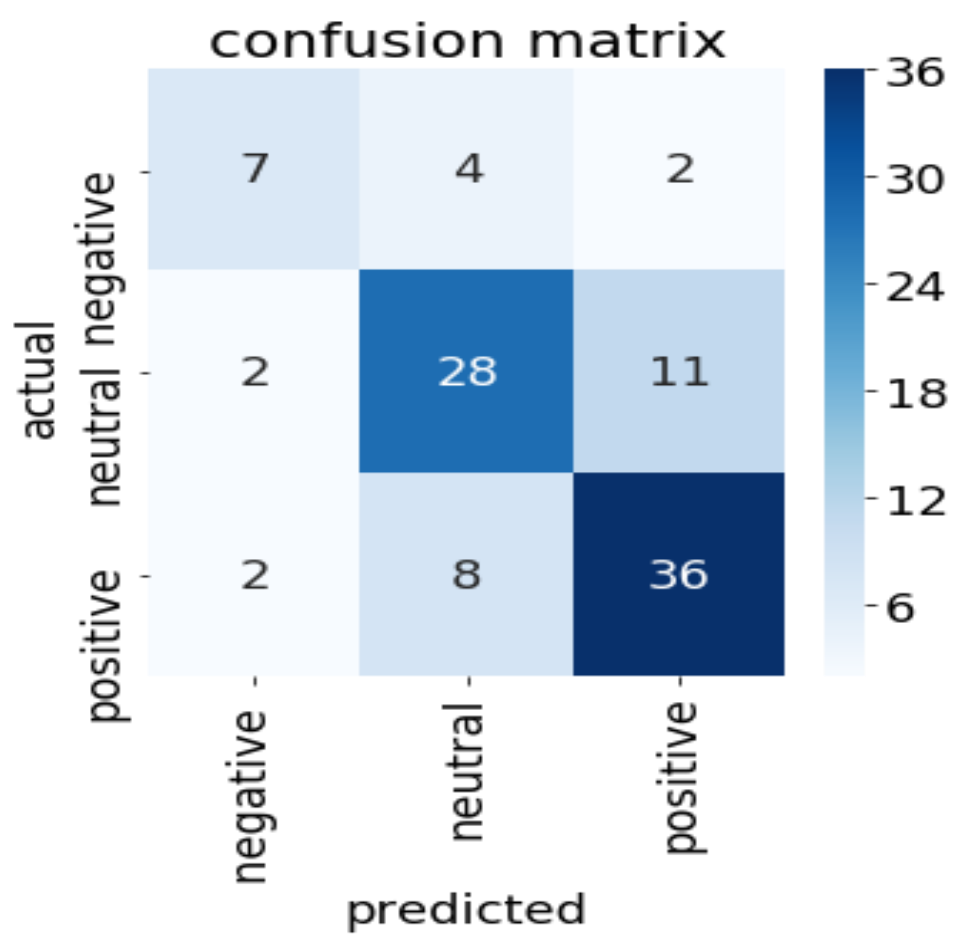

Fig 5. Polarity classification confusion

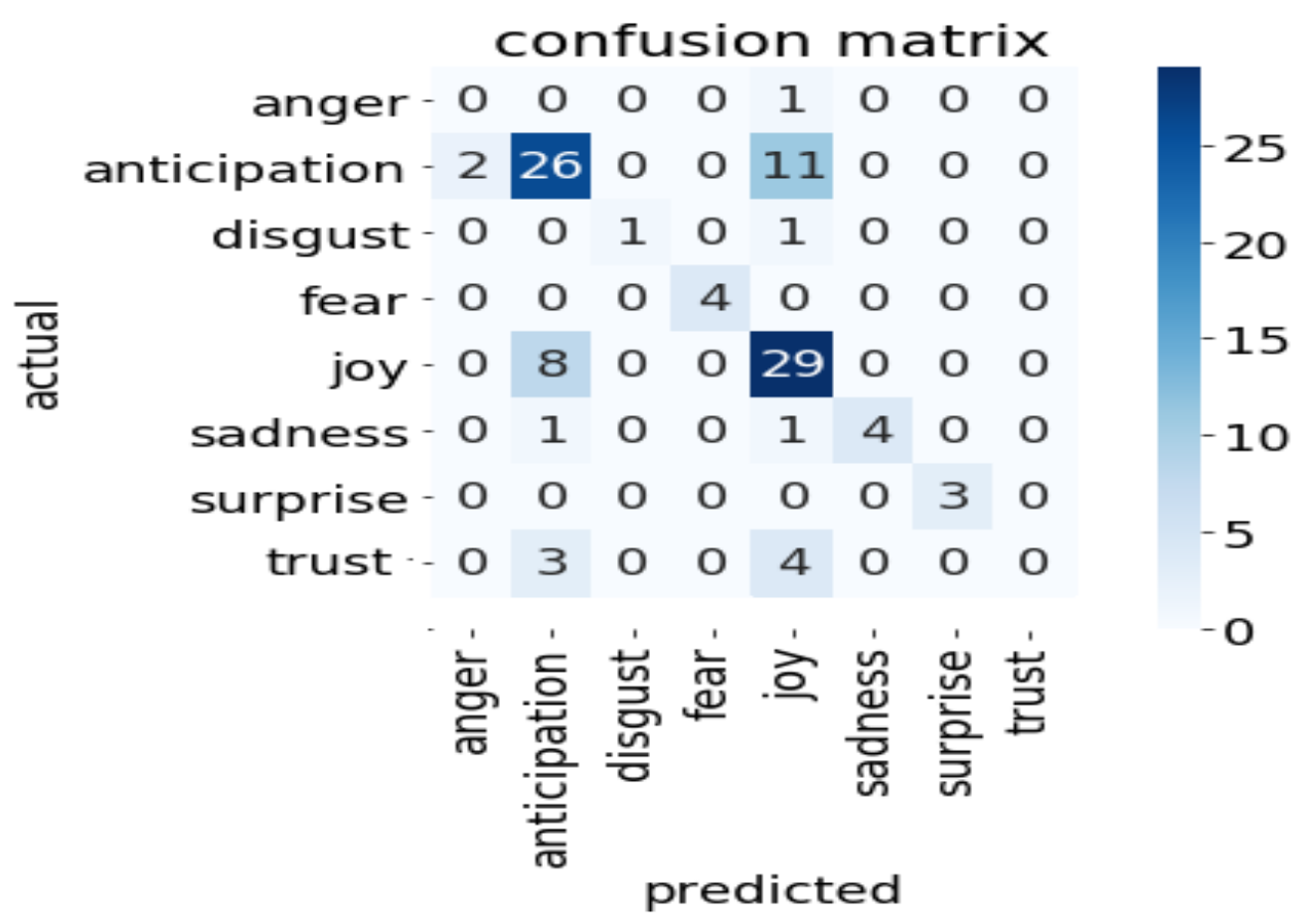

Fig 6. Emotion classification confusion

Fig. 5 and 6 show the confusion matrix for sentiment polarity and emotion classification. We show the actual and the predicted class count. For sentiment polarity, we observe 4 instances of negative classified as neutral, 2 instances classified as positive; 2 instances of neutral classified as negative, 11 instances as positive; 2 instances of positive classified as negative, and 8 instances as neutral. For emotion classification, we observe misclassification of anticipation to joy (11 instances), joy to anticipation (8 instances), trust to joy (4 instances), trust to anticipation (3 instances), anticipation to anger (2 instances), anger to joy (1 instance), disgust to joy (1 instance), sadness to anticipation (1 instance), and sadness to joy (1 instance).

## 8. CONCLUSION & FUTURE WORKS

This work focuses on sentiment polarity and emotion classification for the Nagamese Language and its. In this paper, we build an annotated corpus of 594 manually labelled sentences in Nagamese to apply machine learning techniques. We also built a sentiment dictionary of 1,195 Nagamese words which is used in generating features for the classification purpose. We then predicted sentiment polarity and emotion by building classifier using ML techniques such as Naïve Bayes and SVM. We achieved the highest accuracy of 70.71% for polarity and 67% for emotion on our test dataset. It is intended to create a larger dataset in the future and explore other machine and deep learning models.